\def\year{2020}\relax
\documentclass[letterpaper]{article}
\usepackage{aaai20}
\usepackage{times}
\usepackage{helvet}
\usepackage{courier}
\usepackage[hyphens]{url}
\usepackage{graphicx}
\usepackage{graphicx}
\usepackage{amssymb}
\usepackage{amsmath}
\usepackage{multirow}
\usepackage{bm}
\usepackage{bbm}
\usepackage[ruled]{algorithm2e}
\DeclareMathOperator*{\argmin}{argmin}

\title{Residual Continual Learning}
\author{Janghyeon Lee,\textsuperscript{\rm 1} Donggyu Joo,\textsuperscript{\rm 1} Hyeong Gwon Hong,\textsuperscript{\rm 2} Junmo Kim\textsuperscript{\rm 1}\\
\textsuperscript{\rm 1}School of Electrical Engineering, KAIST\\
\textsuperscript{\rm 2}Graduate School of AI, KAIST\\
\{wkdgus9305, jdg105, honggudrnjs, junmo.kim\}@kaist.ac.kr
}
 
\begin{document}

\maketitle

\begin{abstract}
We propose a novel continual learning method called Residual Continual Learning (ResCL). Our method can prevent the catastrophic forgetting phenomenon in sequential learning of multiple tasks, without any source task information except the original network. ResCL reparameterizes network parameters by linearly combining each layer of the original network and a fine-tuned network; therefore, the size of the network does not increase at all. To apply the proposed method to general convolutional neural networks, the effects of batch normalization layers are also considered. By utilizing residual-learning-like reparameterization and a special weight decay loss, the trade-off between source and target performance is effectively controlled. The proposed method exhibits state-of-the-art performance in various continual learning scenarios.
\end{abstract}

\section{Introduction}

Deep learning with artificial neural networks is now one of the most powerful artificial intelligence technologies. It exhibits state-of-the-art performance in various machine learning fields such as computer vision~\cite{he2016deep}, natural language processing~\cite{wu2016google}, and reinforcement learning~\cite{silver2017mastering}. However, it requires large amounts of training data and time to train such deep networks as network structure becomes complicated. To alleviate this difficulty, transfer learning methods such as fine-tuning~\cite{yosinski2014transferable} are used to utilize source task knowledge and to boost training for target tasks.

As transfer learning methods consider only target task performance during training, most of source task performance is lost as a side effect called the catastrophic forgetting phenomenon~\cite{french1999catastrophic,mccloskey1989catastrophic}. This is a serious problem if high performance is required for both source and target tasks. Continual learning methods should be adopted to resolve this problem.

Our main goal is to achieve good target task performance while maintaining source task performance. Specifically, we focus on image classification tasks with Convolutional Neural Networks (CNNs). Moreover, we impose two practical conditions for the problem.
First, we assume that no source task information is available during target task training.
In many real-world applications, source data are often too heavy to handle or do not have a public license.
If they are available, joint training of source and target data would be a better solution. %, therefore it would be a more appropriate setting
Further, we also assume that not only source data but also any other forms containing source task information are not available.
Recent studies on continual learning do not use the source data directly, but they often refer to parts of the information about source data, for example in forms of generative adversarial networks~\cite{shin2017continual} or Fisher information matrices~\cite{kirkpatrick2017overcoming,ritter2018online}, which somewhat dilutes the original purpose of continual learning.
Second, the size of a network should not increase. Without this condition, a network can be expanded while keeping the entire original network,~\textit{e.g.},~\cite{terekhov2015knowledge,rusu2016progressive}.
Such network expansion methods show good performance in both source and target tasks, but it is difficult to use them with deep neural networks in practice because the size of a network becomes heavier as the number of tasks increases.

The main features are as follows.
\begin{itemize}
\item Residual-learning-like reparameterization allows continual learning, and a simple decay loss controls the trade-off between source and target performance.
\item No information about source tasks is needed, except the original source network.
\item The size of a network does not increase at all for inference (except last task-specific linear classifiers).
\item The proposed method can be applied to general CNNs including Batch Normalization (BN)~\cite{ioffe2015batch} layers in a natural way.
\item We propose two fair measures for comparing different continual learning methods, \textit{maximum achievable average accuracy} for an ideal measure and \textit{source accuracy at required target accuracy} for a practical measure.
\end{itemize}

\section{Related Work}

Learning without Forgetting (LwF)~\cite{li2018learning} is a simple but effective continual learning method. An LwF loss restricts source task outputs of a new network to be close to those of a source network using only target task data. In that work, softmax layers with high temperatures are used to soften the original output distributions in a sense of knowledge distillation~\cite{hinton2015distilling}.
Our method, Residual Continual Learning (ResCL), includes the LwF method as a special case since we use an LwF loss to train a combined network. ResCL falls back to LwF if training is performed on a source network instead of a combined one with true target labels instead of softened ones.

In Incremental Moment Matching (IMM)~\cite{lee2017overcoming}, the posterior distributions of each task and the combined task are approximated as Gaussian distributions. The moments of the posterior distribution for the combined task are incrementally matched by mean-IMM or mode-IMM. Mean-IMM simply averages the weights of the original and new networks as the minimization of the Kullback--Leibler divergence between the posterior of the combined task and the mixture of each task. Mode-IMM merges two networks with their covariance information to approximate the mode of the mixture of two Gaussian posteriors.
Our method also uses the combination of two networks as IMM does. The difference is that the coefficients of the combination in IMM are determined with a hyperparameter of the mixing ratios of each task in a framework of Bayesian neural networks, and there is no additional training for a combined network.
As neural networks are not linear or convex, we cannot be sure that a combined network works properly without additional training.
In ResCL, weights and combination coefficients are learnable, so we can ensure that a combined network works properly.

\cite{terekhov2015knowledge,rusu2016progressive} can prevent forgetting perfectly by expanding a network while keeping the entire original network, but they increase the size of a network.
\cite{kirkpatrick2017overcoming,ritter2018online} protect source task performance by a quadratic penalty loss where the importance of each weight is measured by the Fisher information matrix. However, source data are required to calculate the Fisher information matrix.
\cite{aljundi2018memory} proposes to measure the importance of a parameter by the magnitude of the gradient, which also requires source data.
\cite{zenke2017continual} also defines a quadratic penalty loss designed with the change in a source task loss over an entire trajectory of parameters during source task training.
\cite{mallya2018packnet} adds multiple tasks to a single network by iterative pruning and re-training with source data.

Basically, our method uses linear combination of filters, which has also been studied for multitask or transfer learning, as in \cite{rebuffi2017learning} and \cite{rosenfeld2018incremental} for example. The main purpose of those methods is to learn multiple networks or parameter sets across multiple tasks but with maximized parameter sharing for efficiency, since they focus on multitask or transfer learning.
Therefore, when those are applied to sequential learning, every single task should retain its own network or parameters.
In contrast, our method aims to reduce catastrophic forgetting in sequential learning where only a single network is allowed to represent whole tasks.
In \cite{rebuffi2017learning}, different BN parameters and $1\times1$ filters are learned for each task but with remaining parameters shared across all the given tasks. So, the proposed solution is only applicable to networks that retain BN by definition. Then, it is unsuitable for cases where BN does not work properly, for example, when minibatch size cannot be set large enough due to memory constraints. However, our solution does not depend on BN or specific network architecture, thus more general.
In \cite{rosenfeld2018incremental}, newly added filters for a target task are learned in the form of linear combination with existing filters of source tasks. However, the coefficients necessary for the linear combination are restricted to binary digits $0$ or $1$ during training.
Furthermore, the coefficients are shared across all layers in a network, which could be suboptimal.
In contrast, our method finds a better solution by learning optimal different real coefficients for each layer and each filter during training.

\begin{figure*}[t]
\centering
\includegraphics[width=1.0\textwidth]{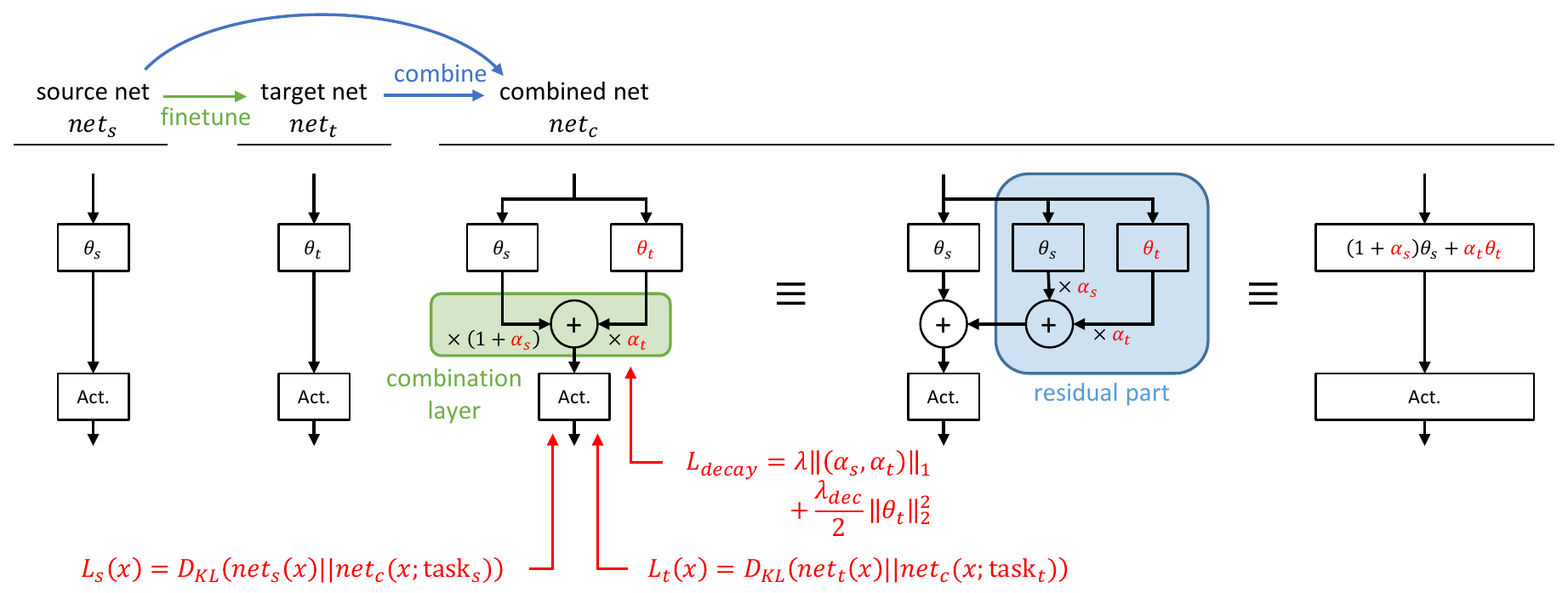}
\caption{An illustration of our method. Learnable parameters are shown in red. We begin with an original network $net_s$, which was trained with source data. First, $net_s$ is fine-tuned with target data to obtain $net_t$. Each linear block in $net_s$ and $net_t$ is combined with a combination layer as in $net_c$. Continual learning on $net_c$ is performed with an LwF loss $L_s$ for preserving performance for the source task and a distillation loss $L_t$ for adapting to the target task. There is also a special decay loss $L_{decay}$, which is the most important loss to prevent forgetting. $D_{KL}(\cdot||\cdot)$ refers to the Kullback--Leibler divergence with a softmax temperature of $2$. Note that each task has its own last task-specific fully connected layer since source and target tasks have different class categories in general. Therefore, $net_c$ has two different outputs: $net_c(\cdot; \textrm{task}_s)$ for a source task and $net_c(\cdot; \textrm{task}_t)$ for a target task.}
\label{fig:method}
\end{figure*}

\section{Method}

Continual learning is essentially to reach a good midpoint between two tasks. A simple idea is linearly combining each layer of source and target networks to obtain a middle network between them, where the source network is the original network that is trained on the source task, and the target network is a fine-tuned network from the original one for the target task. By combining them, we can obtain a network that lies between the source and target task solutions. This basic idea is similar to what IMM~\cite{lee2017overcoming} does, and we also start from here.

However, the performance of a linearly combined network is not guaranteed, as neural networks are not linear or convex. Therefore, after two networks are combined, we have an additional training phase for the combined network to ensure that it will work properly. Because an additional training can hurt the source knowledge, the original weights should be freezed in the combined network. In this paper, we often call the source network as the original network.

\subsection{Linear Combination of Two Layers}

Suppose that we want to combine two fully connected layers whose weight matrices are $\bm{W_s} \in \mathbb{R}^{C_o \times C_i}$ and $\bm{W_t} \in \mathbb{R}^{C_o \times C_i}$. For an input $\bm{x} \in \mathbb{R}^{C_i}$, our combination layer simply combines two outputs linearly with the combination parameters $\bm{\alpha_s} \in \mathbb{R}^{C_o}$ and $\bm{\alpha_t} \in \mathbb{R}^{C_o}$ as follows:
\begin{equation}
\left(\bm{\mathbbm{1}}_{C_o}+\bm{\alpha_s}\right) \circ \left(\bm{W_s x}\right) + \bm{\alpha_t} \circ \left(\bm{W_t x}\right),
\label{eq:combine_fc}
\end{equation}
where $\bm{\mathbbm{1}}_{C_o}$ is a $C_o$-dimensional vector with all ones, and $\circ$ denotes element-wise multiplication. The biases are omitted for brevity. Note that the combination parameters $\bm{\alpha_s}$ and $\bm{\alpha_t}$ are vectors with the dimension $C_o$ and not scalars; thus, the combination layer can set a different importance for each feature. Moreover, $\bm{\alpha_s}$ and $\bm{\alpha_t}$ do not share their values to allow the combination layer to freely manipulate two features. Each combination layer in a network has different values of $\bm{\alpha_s}$ and $\bm{\alpha_t}$ for the same reason.

In the ResCL framework, $\bm{W_s}$ is a weight of the source network, and $\bm{W_t}$ is that of the target network fine-tuned to the target task from the source network. $\bm{W_t}$ is additionally trained to refine its features to be combined well with $\bm{W_s}$, whereas $\bm{W_s}$ is fixed to prevent catastrophic forgetting. Moreover, the combination parameters are also learned by backpropagation to optimally mix two features. Therefore, the learnable parameters in the combined network are $\bm{W_t}$ and $\bm{\alpha} = \left(\bm{\alpha_s}, \bm{\alpha_t}\right)$.

One can easily find that the two fully connected layers and combination layer can be equivalently expressed as one fully connected layer whose weight is
\begin{equation}
\left(\left(\bm{\mathbbm{1}}_{C_o} + \bm{\alpha_s}\right) \bm{\mathbbm{1}}^T_{C_i}\right) \circ \bm{W_s} + \left(\bm{\alpha_t} \bm{\mathbbm{1}}^T_{C_i}\right) \circ \bm{W_t}
\label{eq:combine_fc2}
\end{equation}
owing to their linearity. This is the reason why we can call our method a type of reparameterization; thus, the size of a network does not increase for inference once training is finished. Any nonlinear layer such as sigmoid or ReLU should not be included in the combination, as in Fig.~\ref{fig:method} and Fig.~\ref{fig:arch}. Otherwise, those layers cannot be merged into one layer; then the network size increases as the number of tasks increases.

\subsection{Training}

There are two Kullback--Leibler divergence losses for training of the combined network: one for maintaining the source performance and the other for solving the target task. For the former loss, we adopt an LwF loss~\cite{li2018learning}, which preserves the source information well. That is, the outputs for the source task are constrained to be similar to those of the original network. Softmax layers with a temperature of $2$ are also used. For the latter loss, the distillation loss~\cite{hinton2015distilling} from the fine-tuned network with a temperature of $2$ is used for better generalization and as a natural counterpart of the LwF loss. Therefore, the outputs for the target task are constrained to be similar to those of the fine-tuned network.

One might be wondering why the combination coefficient of $\bm{W_s x}$ is parameterized to $\bm{\mathbbm{1}} + \bm{\alpha_s}$ instead of simply $\bm{\alpha_s}$ in Equation~\ref{eq:combine_fc}.
Since a weight decay loss is widely used as a regularization term to improve generalization of neural networks~\cite{krogh1992simple}, we will also use a weight decay loss for not only $\bm{W_t}$ but also $\bm{\alpha}$. If we use simply $\bm{\alpha_s}$ for the combination coefficient of $\bm{W_s x}$, then the combined output is
\begin{equation}
\bm{\alpha_s} \circ \left(\bm{W_s x}\right) + \bm{\alpha_t} \circ \left(\bm{W_t x}\right) .
\label{eq:combine_decay}
\end{equation}
Now, the source information is lost by the decay loss for $\bm{\alpha_s}$, as it causes the coefficient of $\bm{W_s x}$ to be close to zero. Thus, it is not a good idea to naively decay the combination coefficients; therefore, we carefully design the weight decay loss of combination layers for continual learning. In ResCL, we set the destination of the decay loss to the original network, not a zero-weight network. This can be done by parameterizing the first coefficient of the combination to $\bm{\mathbbm{1}}+\bm{\alpha_s}$ rather than just $\bm{\alpha_s}$, as in Equation~\ref{eq:combine_fc}. With this modification, the decay loss tends to protect the original weights against the target distillation loss. We also experimentally found that the $L_1$ decay loss for $\bm{\alpha}$ is slightly better than $L_2$.

Although it seems to be a very simple reparameterization, it is the key feature that allows continual learning with a significant performance improvement. Actually, this idea is very similar to that of residual learning~\cite{he2016deep}. Residual learning tries to learn the residual of the identity mapping by reformulating a desirable mapping $h(x)$ to $f(x) + x$, where $f(x)$ is a learnable residual function. If the identity mapping is desirable, this can be easily learned by decaying the weights of $f(x)$ to zeros. Similarly, ResCL tries to learn the residual of the source layer by reparameterizing $\bm{W x}$ to $\bm{W_s x} + \bm{\alpha_s} \circ \left( \bm{W_s x} \right) + \bm{\alpha_t} \circ \left( \bm{W_t x} \right)$. If returning to the original source network is desirable to recover forgetting, it can be done easily with a decay loss $\lambda || ( \bm{\alpha_s}, \bm{\alpha_t} ) ||$. 

As learned residual functions in a residual network tend to have small responses~\cite{he2016deep}, if altering some feature is not very helpful for a target task, the decay loss for $\bm{\alpha}$ will settle that feature near the original feature. Only the necessary features for solving the target task will have large deviations from the original features, and the importance of each feature is automatically learned by the decay loss and implicitly controlled by the trade-off hyperparameter $\lambda$ in Algorithm~\ref{algorithm}. As this residual-learning-like reparameterization and the decay loss on $\bm{\alpha}$ play a very important role in our method, we call the proposed continual learning method as Residual Continual Learning. The entire procedure of the proposed ResCL method is summarized in Fig.~\ref{fig:method} and Algorithm~\ref{algorithm}.

\begin{algorithm}
\footnotesize
\SetKwInOut{Input}{Input}
\Input{$net_s\left(\cdot; \bm{\theta_s}^*\right)$   \hfill    // given source network}
\Input{$\lambda$   \hfill    // source--target trade-off hyperparameter}
\Input{$\left(\bm{X_t}, \bm{Y_t}\right)$            \hfill   // training data of target task}
$net_t\left(\cdot; \bm{\theta_t}\right) \gets net_s\left(\cdot; \bm{\theta_s}^*\right);$ \hfill // init $net_t$ as a copy of $net_s$\\
$\bm{\theta_t}^* \gets \argmin_{\bm{\theta_t}} D_{KL}\left(\bm{Y_t}||net_t\left(\bm{X_t}; \bm{\theta_t}\right)\right) + \frac{1}{2} \lambda_{dec} ||\bm{\theta_t}||_2^2;$\\ \hfill // fine-tuning from the source network\\
$\bm{\hat{Y_s}} \gets net_s\left(\bm{X_t}; \bm{\theta_s}^*\right);$ \hfill // source network outputs for LwF\\
$\bm{\hat{Y_t}} \gets net_t\left(\bm{X_t}; \bm{\theta_t}^*\right);$ \hfill // fine-tuned net outputs for distillation\\
$\left(\bm{\alpha_s}, \bm{\alpha_t}\right) \gets \left(-1/2\cdot\bm{\mathbbm{1}}, 1/2\cdot\bm{\mathbbm{1}}\right);$ \hfill // init combination params\\
$\bm{\theta_t} \gets \bm{\theta_t}^*;$ \hfill // init $\bm{\theta_t}$ as fine-tuned weight $\bm{\theta_t}^*$\\
$net_c\left(\cdot; \left(\bm{\alpha_s}, \bm{\theta_s}^*, \bm{\alpha_t}, \bm{\theta_t}\right), \textrm{task}=\cdot\right)$\\
\quad$\gets \textsc{Combine}\left(\bm{\alpha_s}, net_s\left(\cdot; \bm{\theta_s}^*\right), \bm{\alpha_t}, net_t\left(\cdot; \bm{\theta_t}\right)\right);$\\ \hfill // init $net_c$ as in Fig.~\ref{fig:method} and Fig.~\ref{fig:arch}\\
$\left(\bm{\alpha_s}^*, \bm{\alpha_t}^*, \bm{\theta_t}^{**}\right) \gets \argmin_{\bm{\alpha_s}, \bm{\alpha_t}, \bm{\theta_t}} \{$\\
\quad$D_{KL}(\bm{\hat{Y_s}}||net_c\left(\bm{X_t}; \left(\bm{\alpha_s}, \bm{\theta_s}^*, \bm{\alpha_t}, \bm{\theta_t}\right), \textrm{task}=s\right))$\\
\quad$+ D_{KL}(\bm{\hat{Y_t}}||net_c\left(\bm{X_t}; \left(\bm{\alpha_s}, \bm{\theta_s}^*, \bm{\alpha_t}, \bm{\theta_t}\right), \textrm{task}=t\right))$\\
\quad$+ \lambda ||\left(\bm{\alpha_s}, \bm{\alpha_t}\right)||_1 + \frac{1}{2} \lambda_{dec} ||\bm{\theta_t}||_2^2\};$ \hfill // train combined net \\
\SetKwInOut{Output}{Output}
\Output{$net_c\left(\cdot; \left(\bm{\alpha_s}^*, \bm{\theta_s}^*, \bm{\alpha_t}^*, \bm{\theta_t}^{**}\right), \textrm{task}=\cdot\right)$}
\caption{Residual Continual Learning}
\label{algorithm}
\end{algorithm}

\subsection{Convolution and Batch Normalization}

An extension of the proposed framework to convolutional layers is straightforward. Let $\bm{W_s}, \bm{W_t} \in \mathbb{R}^{C_o \times C_i \times H_k \times W_k}$ be the weight tensors of two convolutional layers and $\bm{\alpha_s}, \bm{\alpha_t} \in \mathbb{R}^{C_o}$ be the corresponding combination parameters.
Note that combination parameters are shared across the spatial dimension, in order to take advantage of the structure of CNNs.
Then, the two outputs of the convolutional layers are combined in the same manner as Equation~\ref{eq:combine_fc} with convolutions instead of matrix multiplications.

The case of convolutional layers with BN~\cite{ioffe2015batch} must also be considered since BN is widely used in modern CNNs~\cite{BMVC2016_87,huang2017densely,chollet2017xception}. A BN layer is quite tricky, as it has two different functions depending on its phase of training and inference. BN normalizes an input with the statistics of the current minibatch in the training phase, whereas the population statistics, which are not learned by gradient descent, are used for inference. If training and test data originate from the same task, it is not a significant issue because the two statistics would be very similar. However, it is problematic in continual learning since we are dealing with multiple different tasks whose distributions are not the same in general.

\begin{figure}
\begin{center}
\includegraphics[width=1.0\columnwidth]{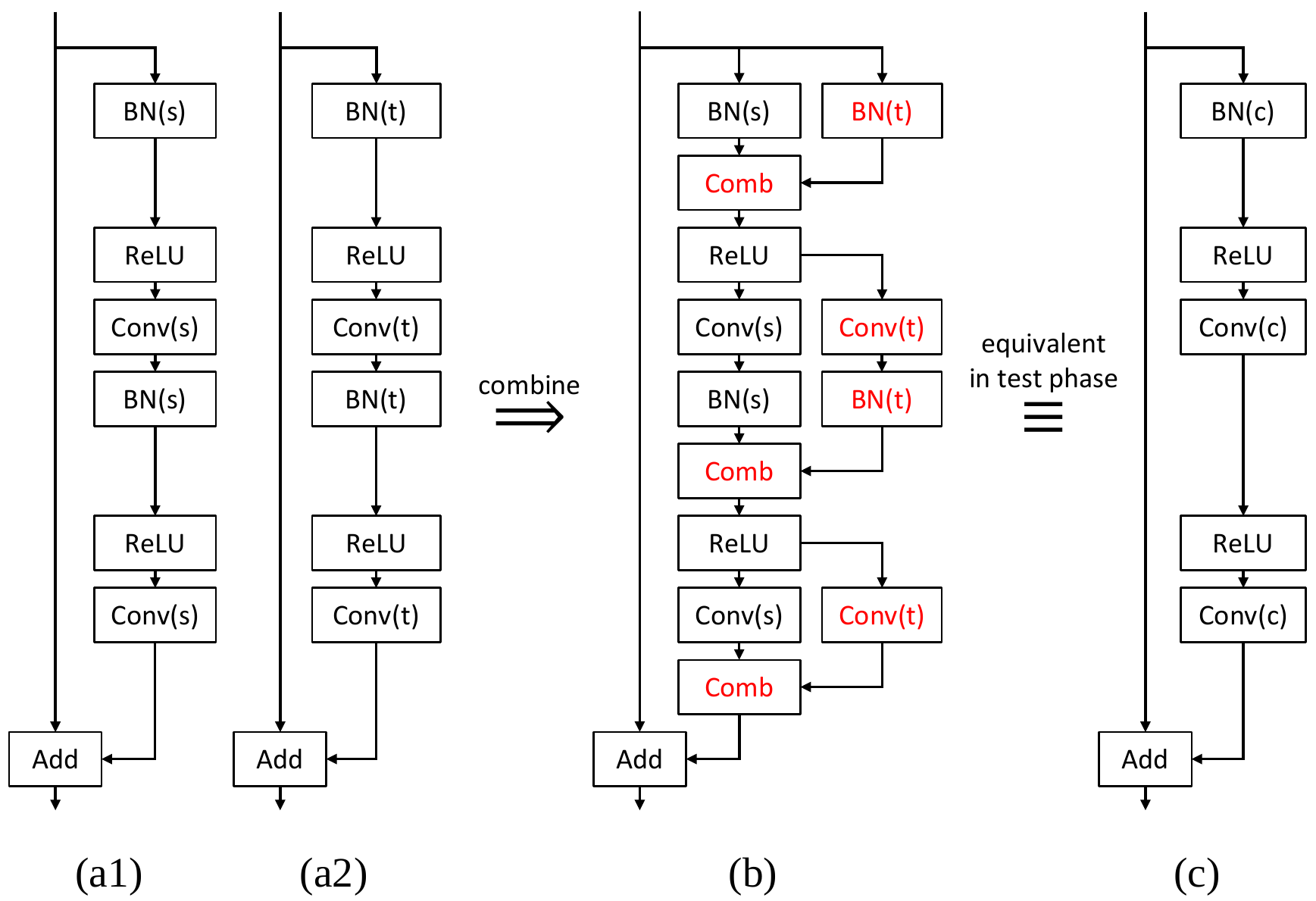}
\end{center}
\caption{Combination of source and target pre-activation residual units~\cite{he2016identity}. ``Comb'' represents a combination layer. Two paths are combined by a combination layer before every nonlinearity. Learnable layers are shown in red. In the inference phase, the combined network (c), which is equivalent to (b) and has the same network size as (a1) and (a2), is used.}
\label{fig:arch}
\end{figure}

Our method can be applied with BN layers in a natural way. We do not need to worry about changes in the distribution, as each subnetwork has its own BN layer for its own task. Specifically, the original BN layer (BN(s) in Fig.~\ref{fig:arch}) should use its population statistics of the source task during both the additional training and test phases. Otherwise, some of the source knowledge is lost, as the combined network cannot see and make use of the original statistics during additional training. The two BN and two convolutional layers with the combination layer can also be merged into one equivalent convolutional layer after training since a BN layer is a deterministic linear layer in the inference phase and convolution is also a linear operation.

\begin{table*}[t]
\centering
\caption{Maximum achievable average accuracies[\%] for each method. Means and standard deviations of four runs. The optimal trade-off hyperparameters are in parentheses.}\smallskip
\resizebox{1.0\textwidth}{!}{
\begin{tabular}{|l||c||c|c|c|}
\hline
Task & (Hyperparam.) & CIFAR-10 $\rightarrow$ CIFAR-100 & CIFAR-100 $\rightarrow$ CIFAR-10 & CIFAR-10 $\rightarrow$ SVHN \\
\hline\hline
Joint Training & $-$ & $80.02\pm0.28$ & $79.63\pm0.06$ & $93.72\pm0.10$ \\
\hline
Last-layer Fine-tuning & $-$ & $61.09\pm0.34$ & $69.71\pm0.33$ & $66.46\pm0.53$ \\
Fine-tuning & $-$ & $62.37\pm1.06$ & $49.90\pm0.37$ & $54.23\pm0.40$ \\
LwF~\cite{li2018learning} & $\lambda$ & $77.67\pm0.10 (2^0)$ & $76.18\pm0.16 (2^1)$ & $68.90\pm0.89 (2^0)$ \\
Mean-IMM~\cite{lee2017overcoming} & $\alpha_1/\alpha_2$ & $73.98\pm0.30 (2^{-9})$ & $76.41\pm0.19 (2^{-10})$ & $79.91\pm1.47 (2^0)$ \\
ResCL (Ours) & $\lambda/10^{-4}$ & $\textbf{78.80}\pm0.17 (2^0)$ & $\textbf{77.13}\pm0.12 (2^{-4})$ & $\textbf{89.49}\pm0.32 (2^5)$ \\
\hline
\end{tabular}
\label{tab:result1-tune}
}
\end{table*}

\begin{table*}[t]
\centering
\caption{Source and target accuracies[\%] for each method.}\smallskip
\renewcommand{\tabcolsep}{5pt}
\begin{tabular}{|l||c|c|c|c|c|c|}
\hline
\multirow{2}{*}{Task} & \multicolumn{2}{c|}{CIFAR-10 $\rightarrow$ CIFAR-100} & \multicolumn{2}{c|}{CIFAR-100 $\rightarrow$ CIFAR-10} & \multicolumn{2}{c|}{CIFAR-10 $\rightarrow$ SVHN} \\
\cline{2-7}
 & source & target & source & target & source & target \\
\hline\hline
Joint Training & $91.70\pm0.19$ & $68.34\pm0.40$ & $68.75\pm0.03$ & $90.51\pm0.12$ & $91.93\pm0.17$ & $95.51\pm0.11$ \\
\hline
Last-layer Fine-tuning & $91.80\pm0.14$ & $30.39\pm0.62$ & $66.98\pm0.25$ & $72.43\pm0.48$ & $91.89\pm0.14$ & $41.11\pm1.06$ \\
Fine-tuning & $56.01\pm2.41$ & $68.74\pm0.46$ & $07.14\pm0.62$ & $92.67\pm0.19$ & $12.49\pm0.84$ & $95.96\pm0.05$ \\
LwF~\cite{li2018learning} & $87.81\pm0.26$ & $67.53\pm0.21$ & $63.91\pm0.14$ & $88.45\pm0.19$ & $43.12\pm1.87$ & $94.68\pm0.14$ \\
Mean-IMM~\cite{lee2017overcoming} & $91.24\pm0.18$ & $56.72\pm0.57$ & $67.42\pm0.17$ & $85.41\pm0.26$ & $81.29\pm2.03$ & $78.53\pm1.53$ \\
ResCL (Ours) & $89.48\pm0.04$ & $68.13\pm0.32$ & $66.84\pm0.39$ & $87.41\pm0.32$ & $88.66\pm0.56$ & $90.32\pm0.23$ \\
\hline
\end{tabular}
\label{tab:result1-tune-st}
\end{table*}

\section{Experiment}
\label{sec:exp}

\subsection{Maximum Achievable Average Accuracy}

We evaluate our method for sequential learning of image classification tasks and compare it with other methods, including fine-tuning, LwF, and Mean-IMM, that do not refer to any source task information for fair comparisons.
Mode-IMM is not compared in the experiment because it requires the Fisher information matrix, which cannot be obtained without source data.
The source and target tasks are to classify the CIFAR-10, CIFAR-100~\cite{krizhevsky2009learning}, or SVHN~\cite{netzer2011reading} dataset. A pre-activation residual network of $32$ layers without bottlenecks~\cite{he2016identity} is used.

For the CIFAR datasets, data augmentation and hyperparameter settings are the same as those in~\cite{he2016identity}. Training images are horizontally flipped with a probability of $0.5$ and randomly cropped to $32\times32$ from $40\times40$ zero-padded images during training. SGD with a momentum of $0.9$, a minibatch size of $128$, and a weight decay of $\lambda_{dec} = 0.0001$ optimizes networks until $64000$ iterations. Note that this `usual' weight decay loss for $\bm{\theta_t}$ is different from the special decay loss for the combination parameters $\bm{\alpha}$. The learning rate starts from $0.1$ and is multiplied by $0.1$ at $32000$ and $48000$ iterations. The He's initialization method~\cite{he2015delving} is used to initialize source networks. Combination parameters $\left(\bm{\alpha_s}, \bm{\alpha_t}\right)$ in ResCL are initialized to $\left(-1/2\cdot\bm{\mathbbm{1}}, 1/2\cdot\bm{\mathbbm{1}}\right)$ in order to balance the original and new features at the early stage of training. For the SVHN dataset, all settings are the same as above, but the training data are not augmented.

We evaluate each method by the average accuracy, which is the average of the source and target accuracies, for three sequential learning scenarios: CIFAR-10 $\rightarrow$ CIFAR-100, CIFAR-100 $\rightarrow$ CIFAR-10, and CIFAR-10 $\rightarrow$ SVHN. Since the source and target tasks have different class categories, each task has its own last task-specific fully connected layer. In LwF, the last layer of target task is trained first with the other weights freezed (warm-up step in \cite{li2018learning}), as in the original paper. For a fair comparison, all other methods also start with this last-layer fine-tuning step. Mean-IMM matches the moments of the last-layer fine-tuning model and LwF model, as in \cite{lee2017overcoming}. ResCL combines two paths before every nonlinearity, as in Fig.~\ref{fig:arch}. The last layer for the target task is not reparameterized because there are no layers for the target task in the original network.

All continual learning methods should control the trade-off between source and target performance by their own trade-off hyperparameters. Since each method has different approaches and different trade-off hyperparameter definitions, it is not a fair comparison to use just one specific trade-off hyperparameter setting. Here, we propose to use a fair measure over different continual learning methods, which does not depend on hyperparameter definitions, \textit{maximum achievable average accuracy}, where the average is taken over all source tasks a model has learned so far and the current target task.
We search the optimal trade-off hyperparameters over $\{2^0, 2^{\pm1}, ..., 2^{\pm10}\}$ for each method to obtain the maximum achievable average accuracies, where the default hyperparameter is $2^0$, and a larger hyperparameter means that the source performance is more strongly protected, for all methods. By this experimental setting, we can obtain the true capacity of each method.
The results are summarized in Tables~\ref{tab:result1-tune} and~\ref{tab:result1-tune-st}. The performance of the joint training method is provided as an upper bound.

LwF works in a small range of trade-off hyperparameters since it directly changes the magnitude of a cross-entropy loss. As the hyperparameter moves away from its default value, the balance between the source and target losses is quickly broken. The optimal trade-off hyperparameters are almost the same as the default value $2^0$ for LwF. Our method can use a wider range of the hyperparameter than LwF by changing the multiplier of the decay loss instead of the cross-entropy loss. It naturally controls how far the reparameterized network is from the original one without any modification of the source and target losses.

\begin{figure}
\begin{center}
\includegraphics[width=1.0\columnwidth]{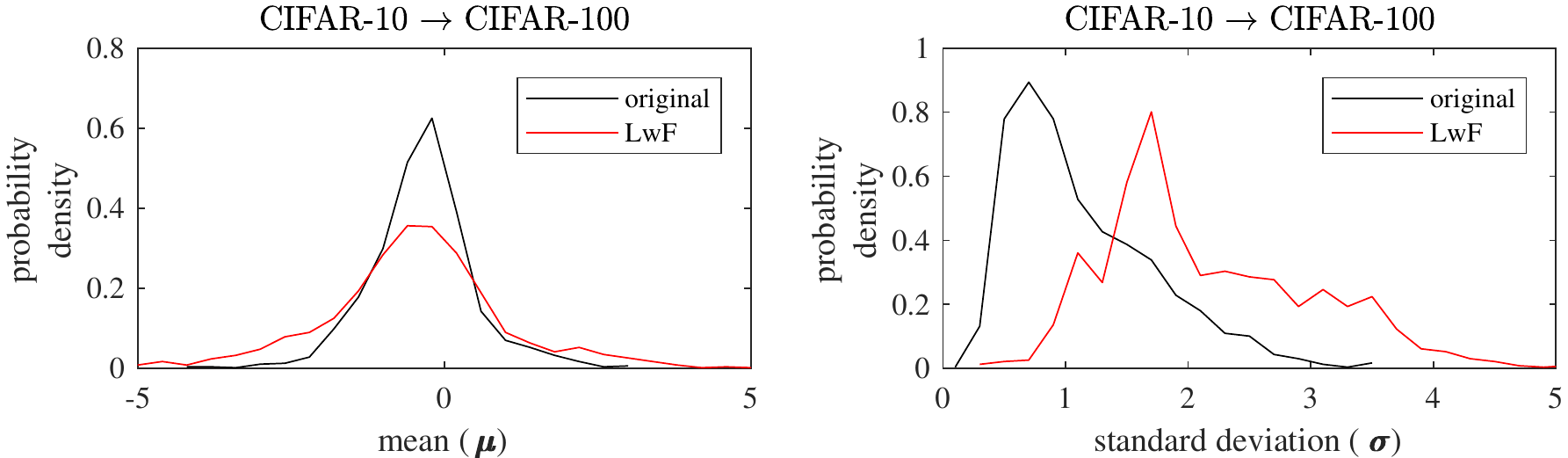}
\end{center}
\caption{The distribution of the statistical parameters ($\bm{\mu}$ \textrm{and} $\bm{\sigma}$) of all BN layers in one specific network. The black line is the distribution of the original network, and the red line is the distribution after the original network is trained on the target task with LwF.}
\label{fig:stat}
\end{figure}

As a wide range of trade-off hyperparameters works effectively in ResCL, we can analyze the meaning of the optimal $\lambda$ in Table~\ref{tab:result1-tune} with the difficulty of each task. First, the CIFAR-10 $\rightarrow$ CIFAR-100 scenario can be thought of as continual learning from an easier task to a harder task because the CIFAR-10 data have $10$ classes to classify and the CIFAR-100 data have $100$ classes. The optimal trade-off hyperparameter is the same as the default value, which means that the default hyperparameter $\lambda = 10^{-4}$ can be used for such coarse-to-fine scenarios. The second scenario, CIFAR-100 $\rightarrow$ CIFAR-10, is the converse of the first scenario. The source task is relatively more difficult than the target task; thus, there is much informative knowledge in the LwF loss, and the original network already has good features for the target task (a target accuracy of $72.43$\% with the last-layer fine-tuning model). Therefore, we can pay less attention to preventing catastrophic forgetting, and the optimal trade-off hyperparameter is small ($1/16$ times the default value).

The CIFAR-10 $\rightarrow$ SVHN case is more challenging. These two tasks are very different; CIFAR-10 images consist of visual objects such as dogs and trucks, whereas the classes of the SVHN dataset are digits. Thus, this scenario is vulnerable to catastrophic forgetting, and other methods perform poorly for the source task. However, ResCL still remarkably maintains the source knowledge well; therefore, it significantly outperforms other methods. The optimal trade-off hyperparameter is large ($32$ times the default value), as the additional training on the target task can easily degrade the source performance.

The statistical parameter ($\bm{\mu}$ \textrm{and} $\bm{\sigma}$) distribution of all BN layers in one specific network is shown in Fig.~\ref{fig:stat} for the LwF method.
The statistics of the two tasks are very different, even though those tasks are similar.
As BN layers contain statistics of the target task only in LwF, we cannot make use of the source population statistics even though they are also an important part of the original network.
In contrast, our method provides BN layers to each task, and further makes use of the source population statistics during training of a combined network.

ResCL can also be used for sequential learning of more than two tasks. In addition to the experiments with the three scenarios, we evaluate our method for the sequential learning of the three tasks to demonstrate the scalability of the proposed method. ResCL still exhibits remarkable performance, as indicated in Table~\ref{tab:result2-tune}. In addition, there is no difficulty in applying the ResCL method to other CNN models or large scale datasets. Table~\ref{tab:result2-tune} summarizes the results for sequential learning from the ILSVRC2012 dataset~\cite{russakovsky2015imagenet} to the Caltech-UCSD Birds-200-2011 dataset~\cite{wah2011caltech} with AlexNet~\cite{krizhevsky2012imagenet} and VGG~\cite{simonyan2014very} architecture.

\begin{table}[t]
\caption{Maximum achievable average accuracies[\%] for each method. The second column represents sequential learning on three tasks.}
\begin{center}
\resizebox{1.0\columnwidth}{!}{
\begin{tabular}{|l||c|c|c|}
\hline
Source task & CIFAR-10 $\rightarrow$ CIFAR-100 & \multicolumn{2}{c|}{ImageNet} \\
\hline
Target task & SVHN & \multicolumn{2}{c|}{CUB} \\
\hline
Network architecture & preResNet & AlexNet & VGG \\
\hline\hline
Last-layer Fine-tuning & $68.76$ & $50.51$ & $64.78$ \\
Fine-tuning & $35.30$ & $46.43$ & $64.54$ \\
LwF~\cite{li2018learning} & $58.44$ & $48.47$ & $68.82$ \\
Mean-IMM~\cite{lee2017overcoming} & $-$ & $52.12$ & $67.88$ \\
ResCL (Ours) & $\textbf{78.53}$ & $\textbf{53.51}$ & $\textbf{68.95}$ \\
\hline
\end{tabular}
}
\end{center}
\label{tab:result2-tune}
\end{table}

\begin{figure}
\begin{center}
\includegraphics[width=1.0\columnwidth]{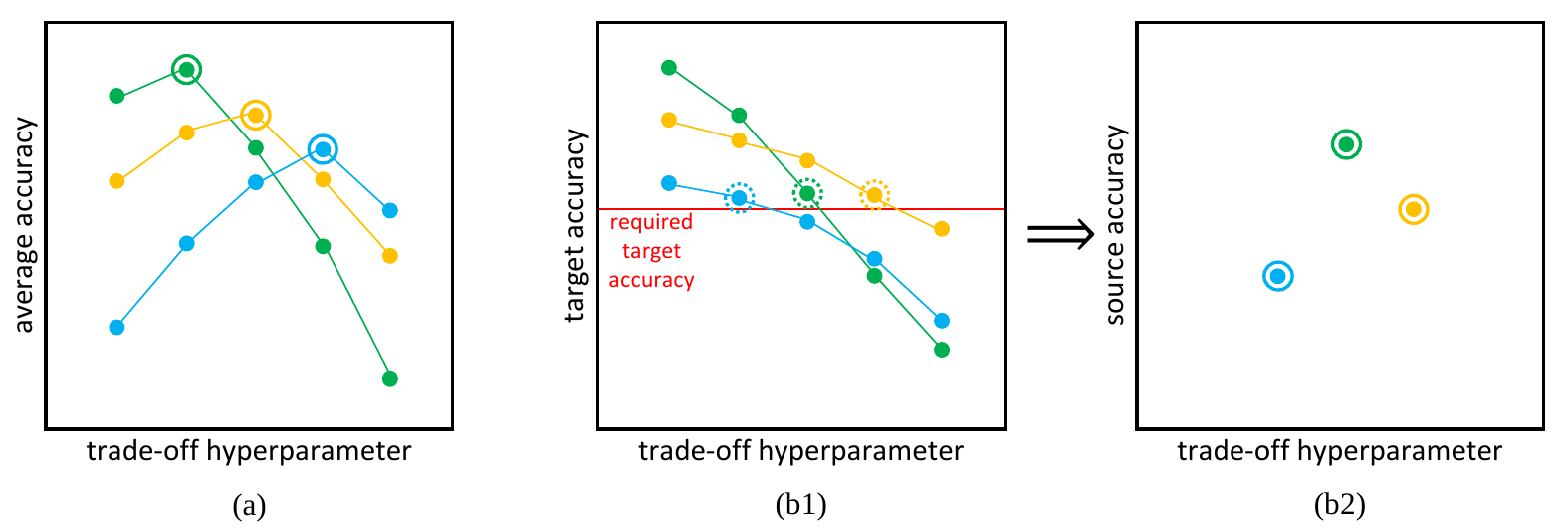}
\end{center}
\caption{As each method has different trade-off hyperparameter definitions, it is not a fair comparison to use just one specific trade-off hyperparameter setting. (a) One of the fair measures is the maximum achievable average accuracy (circled points). (b1) In practice, the trade-off hyperparameter is adjusted using the target validation set until the required target accuracy is reached (dotted circled points). (b2) The models with those hyperparameters are tested once on the source test set.}
\label{fig:eval}
\end{figure}

\subsection{Source Accuracy at Required Target Accuracy}

The maximum achievable average accuracy is a good fair measure that does not depend on trade-off hyperparameter definitions. However, we cannot achieve this maximum average accuracy in practice since there are no available source data for searching the optimal trade-off hyperparameter. This ideal measure gives the true capacity of a continual learning method, but it is not a practical one.

In practical applications, we can set a lower limit of the target accuracy that a model has to achieve. Then, the trade-off hyperparameter can still be effectively adjusted using target validation data only, which are available, until the required target accuracy is reached. After finding the model that meets the required target accuracy, the model is tested once on the source test set (Fig.~\ref{fig:eval}). We set the required target accuracy to $95\%$ of that of the fine-tuning model since the fine-tuning model is trained to solve only the target task well. This evaluation setting gives another fair measure, the \textit{source accuracy at required target accuracy}, which can be determined in practice. The results with this measure are summarized in Tables~\ref{tab:result1-prac} and~\ref{tab:result2-prac}.

\begin{table}[t]
\caption{Source accuracies at required target accuracy[\%] of each method.}
\begin{center}
\resizebox{1.0\columnwidth}{!}{
\begin{tabular}{|l||c|c|c|}
\hline
Source task & CIFAR-10 & CIFAR-100 & CIFAR-10 \\
\hline
Target task & CIFAR-100 & CIFAR-10 & SVHN \\
\hline
Required target accuracy & $65.30$ & $88.04$ & $91.17$ \\
(w.r.t. fine-tuning model) & $(95\%)$ & $(95\%)$ & $(95\%)$ \\
\hline\hline
LwF~\cite{li2018learning} & $89.59$ & $63.91$ & $38.08$ \\
ResCL (Ours) & $\textbf{90.65}$ & $\textbf{68.13}$ & $\textbf{76.83}$ \\
\hline
\end{tabular}}
\end{center}
\label{tab:result1-prac}
\end{table}

\begin{table}[t]
\caption{Source accuracies at required target accuracy[\%] of each method. The second column represents sequential learning on three tasks.}
\begin{center}
\resizebox{1.0\columnwidth}{!}{
\begin{tabular}{|l||c|c|c|}
\hline
Source task & CIFAR-10 $\rightarrow$ CIFAR-100 & \multicolumn{2}{c|}{ImageNet} \\
\hline
Target task & SVHN & \multicolumn{2}{c|}{CUB} \\
\hline
Network architecture & preResNet & AlexNet & VGG \\
\hline
Required target accuracy & $90.07$ & $50.83$ & $67.04$ \\
(w.r.t. fine-tuning model) & $(95\%)$ & $(95\%)$ & $(95\%)$ \\
\hline\hline
LwF~\cite{li2018learning} & $41.84$ & $40.17$ & $66.55$ \\
Mean-IMM~\cite{lee2017overcoming} & $-$ & $52.33$ & $68.48$ \\ 
ResCL (Ours) & $\textbf{53.24}$ & $\textbf{53.91}$ & $\textbf{69.73}$ \\
\hline
\end{tabular}}
\end{center}
\label{tab:result2-prac}
\end{table}

\subsection{Trade-off Hyperparameter $\lambda$ and Combination Parameter $\bm{\alpha}$}

\begin{figure}
\begin{center}
\includegraphics[width=1.0\columnwidth]{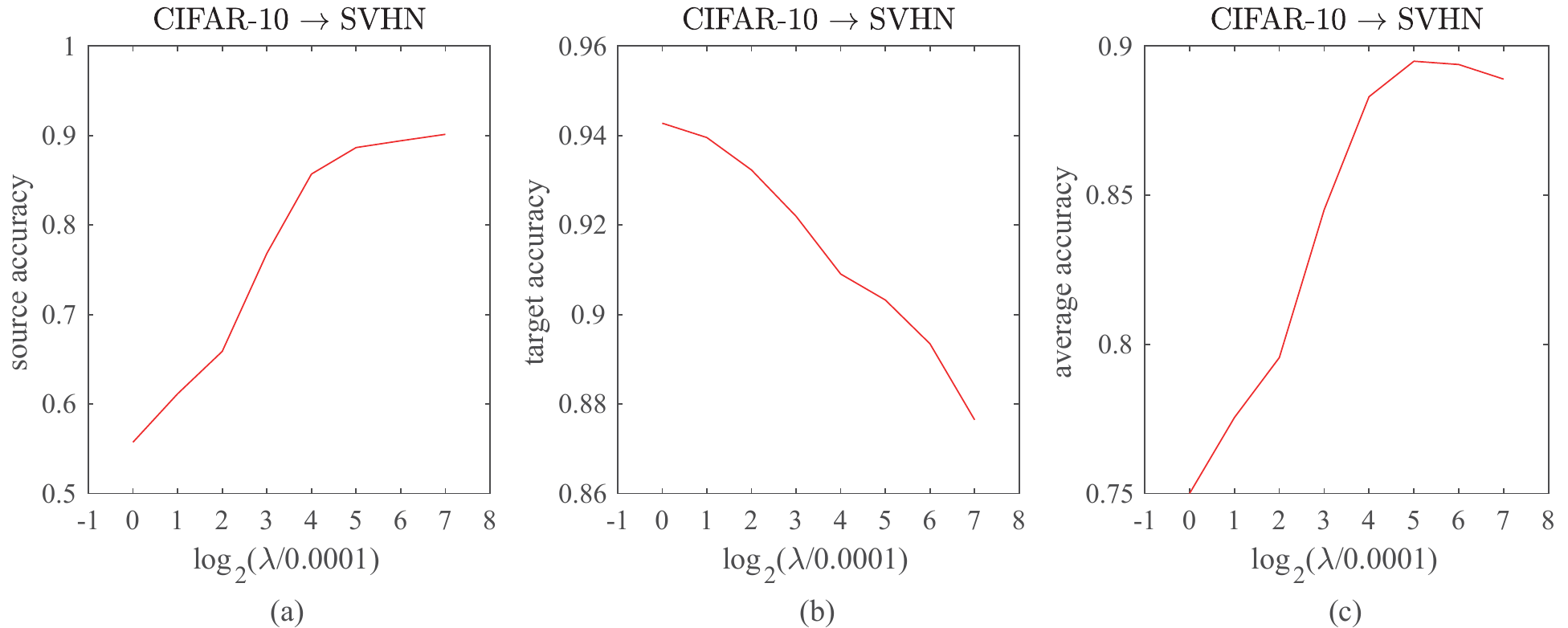}
\end{center}
\caption{Source, target, and average accuracies with respect to the trade-off hyperparameter $\lambda$ in the CIFAR-10 $\rightarrow$ SVHN scenario.}
\label{fig:lambda}
\end{figure}

In this section, we investigate whether the hyperparameter $\lambda$, which is the multiplier of the decay loss for the combination parameter $\bm{\alpha}$, controls the source--target performance trade-off reasonably. For the CIFAR-10 $\rightarrow$ SVHN case, the source, target, and average accuracies with respect to $\lambda$ are shown in Fig.~\ref{fig:lambda}.
As shown in Fig.~\ref{fig:lambda}, $\lambda$ acts as a reasonable trade-off hyperparameter. The source accuracy increases with $\lambda$ and becomes saturated, whereas the target accuracy is a decreasing function of $\lambda$. As a result, the average accuracy graph has a concave shape.

\begin{figure}
\begin{center}
\includegraphics[width=1.0\columnwidth]{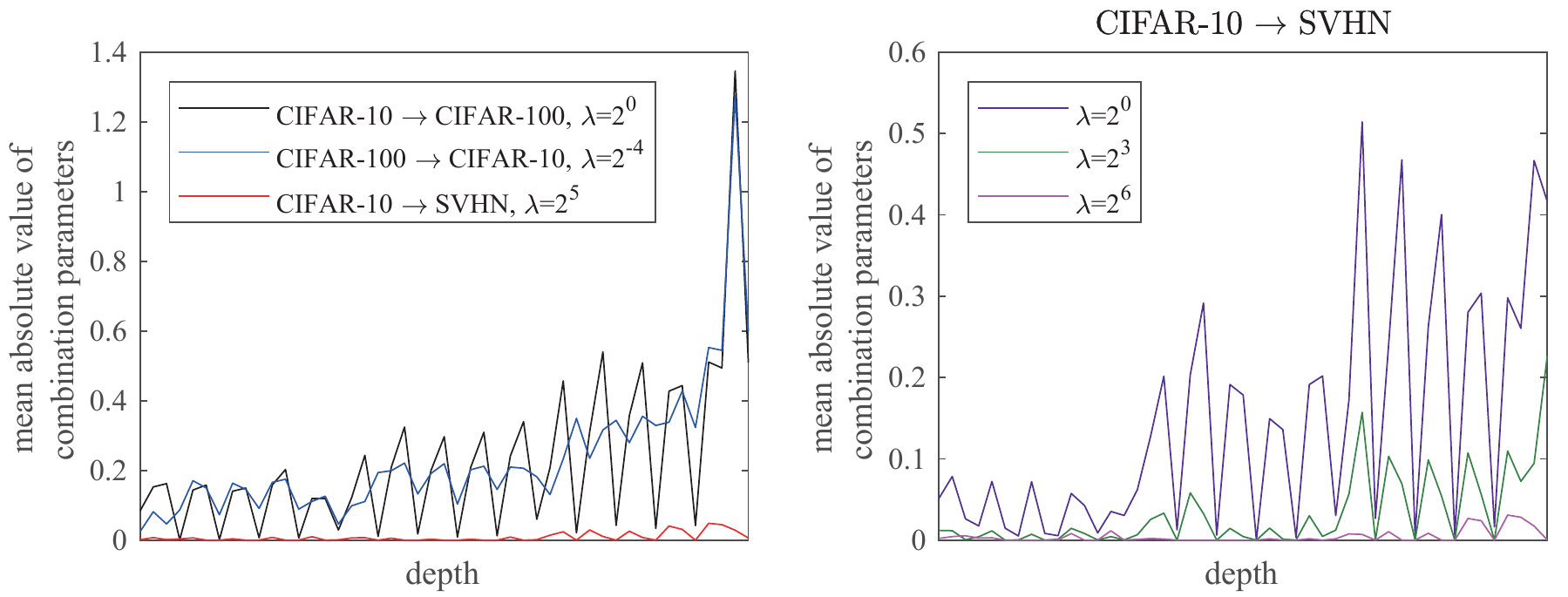}
\end{center}
\caption{Mean absolute value of the elements of the combination parameter $\bm{\alpha}$ with respect to the depth.}
\label{fig:alpha}
\end{figure}

By probing the magnitude of $\bm{\alpha}$, we can observe how much the features was changed to solve the target task. Fig.~\ref{fig:alpha} shows the mean absolute value of the elements of combination parameters with respect to the depth of their layers. For all scenarios in Fig.~\ref{fig:alpha}, the magnitude of the changes tends to increase with the depth. In deep convolutional neural networks, it is known that shallow layers learn basic features such as colors, edges, and corners, whereas deep layers learn class-specific features such as dog faces and bird legs~\cite{zeiler2014visualizing}. Thus, shallow layers do not need to be changed much since their features are already common in both tasks, but deep layers have large deviations from the original ones since the classes of the target task are different from those of the source task.

\section{Conclusion}

We have proposed a novel continual learning method, ResCL, which exhibits the state-of-the-art performance for continual learning of image classification tasks. It prevents catastrophic forgetting, even if the source and target tasks are very different. ResCL can be used in practice, as no information about the source task is required, except the original network, and the size of a network does not increase in the inference phase. Moreover, any general CNN architectures can be adopted since our method is designed to handle convolution and BN layers.

In this study, we limited the scope of the task to sequential learning of image classification with CNNs. However, the ResCL method can be naturally extended to support other types of neural networks, such as recurrent neural networks, since it simply linearly combines the outputs of two layers. We leave the application of the ResCL method to other fields for future work.

\section{Acknowledgment}

This research was supported by the Engineering Research Center Program through the National Research Foundation of Korea (NRF) funded by the Korean Government MSIT (NRF-2018R1A5A1059921).

\bibliography{bib}
\bibliographystyle{aaai}

\end{document}